\documentclass[a4paper]{article}

\usepackage[a4paper]{geometry}

\newif\ifdraft
\draftfalse

\usepackage{ifthen}
\usepackage{natbib}
\usepackage{amsmath,amssymb}
\usepackage{amsthm}
\usepackage{bm}
\usepackage[usenames,dvipsnames,svgnames,table]{xcolor}
\usepackage[hyperindex,
            linktocpage=true,
            colorlinks=true,
            linkcolor=blue,
            urlcolor=blue,
            citecolor=blue,
            anchorcolor=blue
            ]{hyperref}
\usepackage{stmaryrd}
            
\ifdraft
\usepackage[notref,notcite]{showkeys}
\fi

\usepackage{algorithm}
\usepackage{subcaption}
\usepackage{listings}

\usepackage{letltxmacro}
\newcommand*{\SavedLstInline}{}
\LetLtxMacro\SavedLstInline\lstinline
\DeclareRobustCommand*{\lstinline}{
  \ifmmode
    \let\SavedBGroup\bgroup
    \def\bgroup{
      \let\bgroup\SavedBGroup
      \hbox\bgroup
    }
  \fi
  \SavedLstInline
}

\usepackage[capitalize]{cleveref}
\crefname{listing}{Algorithm}{Algorithms}
\Crefname{listing}{Algorithm}{Algorithms}

\usepackage{graphicx}
\usepackage{fancyhdr}

\theoremstyle{definition}

\newcommand{\mynewtheorem}[3]{
\ifthenelse{\equal{#1}{theorem}}{
    \newtheorem{my#1}{#2}
}{
    \newtheorem{my#1}[mytheorem]{#2}
    \ifthenelse{\equal{#3}{}}{
        \crefname{my#1}{#2}{#2s}
    }{
        \crefname{my#1}{#2}{#3}
    }
}
}

\mynewtheorem{theorem}{Theorem}{}
\mynewtheorem{lemma}{Lemma}{}
\mynewtheorem{corollary}{Corollary}{Corollaries}
\mynewtheorem{definition}{Definition}{}
\mynewtheorem{assumption}{Assumption}{}
\mynewtheorem{proposition}{Proposition}{}
\mynewtheorem{remark}{Remark}{}
\mynewtheorem{example}{Example}{}
\mynewtheorem{exercise}{Exercise}{}

\newenvironment{remark}
  {\pushQED{\qed}\myremark}
  {\popQED\endmyremark}

\definecolor{darkred}{rgb}{.5,0,0}
\definecolor{darkgreen}{rgb}{0,.5,0}
\definecolor{darkblue}{rgb}{0,0,.5}
\definecolor{darkorange}{rgb}{.8,.4,0}
\ifdraft
\newcommand{\todo}[1]{\textcolor{darkorange}{(\emph{TODO: #1})}}
\newcommand{\comment}[1]{\textcolor{gray}{(\emph{#1})}}
\newcommand{\warning}[1]{\textcolor{red}{(\emph{WARNING: #1})}}
\newcommand{\quest}[1]{\textcolor{darkgreen}{(\emph{Q: #1})}}

\newcommand{\mh}[1]{\textcolor{darkblue}{(MH: \emph{#1}})}
\else
\newcommand{\todo}[1]{}
\newcommand{\comment}[1]{}
\newcommand{\warning}[1]{}
\newcommand{\quest}[1]{}

\newcommand{\mh}[1]{}
\fi

\newcounter{alphoversetcount}

\newcommand{\resetalph}{\setcounter{alphoversetcount}{0}}

\newenvironment{alphalign*}{
\csname align*\endcsname\resetalph{}
}{
\csname endalign*\endcsname\resetalph{}
}

\newcommand{\eg}{\emph{e.g.}, }

\DeclareMathOperator*{\argmax}{argmax}

\newcommand{\D}{\mathrm{d}}

\newcommand{\floor}[1]{\left\lfloor#1\right\rfloor}

\usepackage{tikz}
\usetikzlibrary{positioning}

\title{Super-Exponential Regret for UCT, AlphaGo and Variants}
 
\author{Laurent Orseau, Rémi Munos}
\date{Google DeepMind \\ \{lorseau,munos\}@google.com}

\begin{document}

\maketitle

\begin{abstract}
We improve the proofs of the lower bounds of \citet{coquelin2007bandit} 
that demonstrate that UCT can have $\exp(\dots\exp(1)\dots)$ regret
(with $\Omega(D)$ exp terms)
on the $D$-chain environment, and that a `polynomial' UCT variant has $\exp_2(\exp_2(D - O(\log D)))$ regret on the same environment ---
the original proofs contain an oversight for rewards bounded in $[0, 1]$, which we fix in the present draft.
We also adapt the proofs to AlphaGo's MCTS and its descendants (\eg AlphaZero, Leela Zero) to also show 
$\exp_2(\exp_2(D - O(\log D)))$ regret.
\end{abstract}

\section{Introduction}

First we present the $D$-chain environment, and then provide lowers bounds for Polynomial UCT, AlphaZero, and UCT.

\section{The $D$-chain environment}

The $D$-chain environment \citep{coquelin2007bandit} is as follows --- see \cref{fig:D-chain}.
Consider a binary tree of depth $D$ with two actions, 1 and 2.
After taking $d<D$ times the action 1 from the root, with 0 reward,
taking the action 2 leads to a terminal state with reward $1-(d+1)/D$.
After taking $D$ times the action 1, the next state is a terminal state with reward 1 --- this is the optimal trajectory.

We call $n_d$ the node reached after taking $d$ times the action 1,
and we call $n_{d'}$ the node reached after taking $d-1$ times the action 1, then the action 2. The node $n_0$ is the root.

\begin{figure}
    \centering

\def\D{6}

\begin{tikzpicture}[level distance=1cm, sibling distance=2cm]

    \tikzstyle{mynode} = [circle, draw, minimum size=0.5cm]

    \node[mynode] (n0) {};
    \foreach \i in {1,...,\D} {
        \pgfmathsetmacro{\j}{int(\i-1)}
        \ifnum \i=2
            \node[above right=0.2cm and 1cm of n\j,minimum size=0.5cm] (n\i) {...};
        \else
            \node[above right=0.2cm and 1cm of n\j,mynode] (n\i) {};
        \fi
        \draw (n\j) -- node[above,midway] {1} (n\i);
        \ifnum \i=3
        \else
            \node [below right=0.2cm and 1cm of n\j,mynode] (branch\i 2) {};
            \draw (n\j) -- node[above,midway] {2} (branch\i 2);
        \fi
    }
    \node[right=0ex of branch12] () {(D-1)/D};
    \node[right=0ex of branch22] () {(D-2)/D};
    \node[right=0ex of branch42] () {2/D};
    \node[right=0ex of branch52] () {1/D};
    \node[right=0ex of branch62] () {0};
    \node[right=0ex of n6] () {1};

    \foreach \i in {1,...,\D} {
        \pgfmathsetmacro{\j}{int(\i-1)}
        
    }

\end{tikzpicture}

    \caption{The $D$-chain environment. Edge labels are actions,
    and node labels are rewards.}
    \label{fig:D-chain}
\end{figure}

\begin{remark}
Since the environment is deterministic, algorithms could take advantage of this by not visiting terminal states more than once, meaning that the lower bounds would not apply to such algorithms.
To make matters more interesting, we can instead assume that there is no terminal state and that the tree is an infinite binary tree.
This forces search algorithms to visit the same nodes multiple times in search of better rewards elsewhere in the tree.
(Making the environment stochastic would substantially complicate the analysis.)
\end{remark}

\section{Polynomial UCT lower bound}

A trajectory is a sequence of states/actions starting from the root and ending in a terminal state.
Let $m_{i,t}$ be the number of trajectories going through node $n_i$, after observing $t$ trajectories.
Note that at the root $n_0$ we have $m_{0,t} = t$.
Define $X_{i,t}$ to be the empirical mean of the rewards obtained
on the $m_{i,t}$ trajectories going through node $n_i$.

\citet{coquelin2007bandit} propose a variant of UCT~\citep{kocsis2006uct} as follows.
Define, at trajectory $t+1$, for $i\in\{d,d'\}$ for each $d< D$,
\begin{align}
    B_{i,t+1} = X_{i,t} + \sqrt{\frac{\sqrt{m_{d-1,t}}}{m_{i,t}}}\,.
\end{align}
If at trajectory $t+1$ the node $n_{d-1}$ is visited,
then the node $n_j$ with $j\in\argmax_{i\in\{d, d'\}} B_{i,t+1}$ is visited,
with tie breaking in favour of $d'$.
The corresponding action is $1$ if $j=d$, or $2$ otherwise.

Let $T+1$ be the first step at which the node $n_D$ (with the maximum reward 1) is reached.
Then all the nodes $n_d$ and none of the nodes $n_{d'}$ are visited on this trajectory,
so for each $d\leq D$ we necessarily have $B_{d,T} \geq B_{d', T}$.
Moreover, due to tie-breaking, all nodes $n_{d'}$ for $d\leq D$ have been visited at least once, and thus $X_{d',T} = 1- d/D$.
Also note that $X_{d,T} \leq 1-(d+1)/D$.
Therefore, for all $d\leq D$,
\begin{align}
    X_{d,T} + \sqrt{\frac{\sqrt{m_{d-1,T}}}{m_{d,T}}}
    \geq 
    X_{d',T} + \sqrt{\frac{\sqrt{m_{d-1,T}}}{m_{d',T}}}\,.
\end{align}
For the rest of the proof, let us make the subscripts $T$ implicit for visual clarity.
It follows that
\begin{align}
    1-\frac{d+1}{D} + \sqrt{\frac{\sqrt{m_{d-1}}}{m_{d}}}
    \geq 
    1-\frac{d}{D} + \sqrt{\frac{\sqrt{m_{d-1}}}{m_{d'}}}
\end{align}
and thus
\begin{align}\label{eq:rec}
    \sqrt{\frac{\sqrt{m_{d-1}}}{m_{d}}}
    \geq \sqrt{\frac{\sqrt{m_{d-1}}}{m_{d'}}} + \frac1D\,.
\end{align}
From this, by dropping the term with $m_{d'}$, we deduce that for all $d\leq D$,
\begin{align}\label{eq:rec_square}
    m_{d-1} \geq \left(\frac{m_d}{D^2}\right)^2\,.
\end{align}

\iffalse
\citet{coquelin2007bandit} then use this relation to assert incorrectly that:
\begin{align}
    m_0 
    &\geq \frac{m_1^2}{D^4}
    \geq \frac{m_2^{2^2}}{D^{4(1+2)}}
    \geq \frac{m_3^{2^3}}{D^{4(1+2+3)}}
    \geq \dots
    \geq \frac{m_{D-1}^{2^{D-1}}}{D^{2D(D-1)}}
    &\text{(incorrect)}
\end{align}
and then conclude
\footnote{This is actually also incorrect because $m_{D-1,T} =1$ only, while we will have $m_{D-1,T+1}=2$ \emph{after} the $T+1$th trajectory. We would also have $m_{D-1,T} =2$ if each trajectory was expanding exactly one non-expanded node as is commonly done in MCTS algorithms, but this is not the definition used in Algorithm 1 from \citet{coquelin2007bandit}.}
by using $m_{D-1} = 2$.
Indeed, the correct recurrent application of \cref{eq:rec_square} gives,
writing $\tilde D = D^2$ for clarity:
\begin{align}
    m_0 
    &\geq \frac{m_1^2}{\tilde D^2} 
    = \frac1{\tilde D^{2^1}}m_1^{2^1}
    \geq \frac1{\tilde D^{2^1}}\left(\frac{m_2^2}{\tilde D^2}\right)^{2^1}
    = \frac1{\tilde D^{2^1+2^2}}m_2^{2^2}
    \geq \frac1{\tilde D^{2^1+2^2}}\left(\frac{m_3^2}{\tilde D^2}\right)^{2^2}
    = \frac1{\tilde D^{2^1+2^2+2^3}}m_3^{2^3} \notag\\
    &\geq \dots
    \geq \frac1{\tilde D^{2^1+2^2+2^3+\dots 2^{D-1}}}m_3^{2^{D-1}}
    \geq \frac1{\tilde D^{2^D}}m_{D-1}^{2^{D-1}} \notag\\
    &= \left(\frac{m_{D-1}}{D^4}\right)^{2^{D-1}}\,.
    \label{eq:rec_square_multi}
\end{align}
But now even using $m_{D-1}= 2$ leads to a vacuous bound.
\subsection{Fixing the analysis for rewards bounded by $D$}

By replacing the rewards of
we obtain the recurrence relation $m_{d-1} \geq m_d^2$, which means the denominator in \cref{eq:rec_square_multi} becomes just 1 and the bound is not vacuous anymore,
leading to a lower bound of $2^{2^{D-1}}$ steps to reach the maximum reward of $D$.

For UCT the regret bound indeed becomes $\Omega(\exp(\dots \exp(1)\dots))$.

However, it is often assumed that rewards should be bounded by 1.
Since the algorithm is not scale-invariant, we need to adapt the analysis.

\subsection{Fixing the analysis for rewards bounded in $[0, 1]$}\label{sec:bounded_rewards}

Fortunately, the original statement of $\Omega(\exp(\exp(D))$ regret is still (mostly) correct.

\else
\begin{remark}
If we replace the rewards $(D-d)/D$ in the environment with just $D-d$, 
then the original claims follow trivially.
Indeed \cref{eq:rec} becomes $m_{d-1}\geq (m_d)^2$, and using
\footnote{Observe that $m_{D-1}=1$ only, which is not sufficient for bootstrapping the sequence.}
$m_{D-2}=2=2^{2^0}$ we deduce that $m_0 \geq 2^{2^{D-2}}$.
However, it is often assumed that rewards should be bounded by 1.
Since the algorithm is not scale-invariant, we need to adapt the analysis.
\end{remark}
\fi

By applying \cref{eq:rec_square} recursively we obtain for all $d < D$,
writing $\tilde D = D^2$ for clarity,
\begin{align}
    m_0 
    &\geq \frac{m_1^2}{\tilde D^2} 
    = \frac1{\tilde D^{2^1}}m_1^{2^1}
    \geq \frac1{\tilde D^{2^1}}\left(\frac{m_2^2}{\tilde D^2}\right)^{2^1}
    = \frac1{\tilde D^{2^1+2^2}}m_2^{2^2}
    \geq \frac1{\tilde D^{2^1+2^2}}\left(\frac{m_3^2}{\tilde D^2}\right)^{2^2}
    = \frac1{\tilde D^{2^1+2^2+2^3}}m_3^{2^3} \notag\\
    &\geq \dots
    \geq \frac1{\tilde D^{2^1+2^2+2^3+\dots 2^d}}m_3^{2^d}
    \geq \frac1{\tilde D^{2^{d+1}}}m_d^{2^d} \notag\\
    &= \left(\frac{m_d}{D^4}\right)^{2^d}\,.
    \label{eq:m0_md}
\end{align}
We want to find a value $\hat d$ for $d$ such that $m_{\hat d}/D^4 \geq 2$.
To do so,
by dropping the $1/D$ term in \cref{eq:rec} and simplifying,
\footnote{In case of iterative expansion of the tree as is often done in practice,
we would have $m_{d-1} = m_d + m_{d'} + 1$ and thus still $m_{d-1} \geq 2 m_d$.}
we deduce that for all $d \leq D$ we have $m_d \leq m_{d'}$,
and thus $m_{d-1}\geq 2m_d$.
Applying this relation recursively gives us that 
\begin{align}\label{eq:mhatd}
    m_{\hat d}\geq 2^1 m_{\hat d+1}\geq 2^2m_{\hat d+2}
    \geq\dots\geq
    2^{D-1-\hat d} m_{D-1}
    = 2^{D-\hat d-1}
\end{align}
using $m_{D-1, T} = 1$.
Now, as said above, we want $m_{\hat d}/D^4 \geq 2$, which is satisfied when
$2^{D-\hat d-1} \geq 2D^4$,
which is satisfied for $\hat d = \floor{D-2-4\log_2 D}$.
Plugging $d=\hat d$ into \cref{eq:m0_md} we obtain
(with $\exp_2(x) = 2^x$)
\begin{align}
    T = m_0 \geq \exp_2(\exp_2(\floor{D-2-4\log_2 D}))\,.
\end{align}
For $D=25$, this gives $T \geq 2^{1024}\geq 10^{100}$, which is intractable,
while a simple breadth-first search in a full complete binary tree of depth 25 would take only $2^{25}$ search steps --- which is well tractable.

\section{AlphaZero lower bound}

In this section we merely adapt the steps for Polynomial UCT
to a different definition fo $B_{i,t}$, and we choose in particular a different
$\hat d$.

The action selection of the `MCTS' algorithm in AlphaGo \citep{silver2016alphago}
and its successors, \eg AlphaGo Zero~\citep{silver2017mastering}, AlphaZero~\citep{silver2017zero} and LeelaChess Zero,
\footnote{\url{https://slides.com/crem/lc0\#/9}}
has the following form for the $D$-chain environment,
for $i\in\{d,d'\}$ for all $d<D$,
\begin{align}
    B_{i, t+1} = Q_{i,t} + c_{\text{puct}}P_i \frac{\sqrt{m_{d-1}}}{m_i+1}
\end{align}
where $c_{\text{puct}}$ is a small positive constant such as 2 or 4,
$P_i$ is the policy weight of action $i$ such that $P_d + P_{d'} = 1$ for all $d\leq D$,
and $Q_{i,t}$ is the ``combined mean action value''.
In what follows we assume that 
(i) $Q_{i,t} = X_{i,t}$ as defined above,
(ii) $P_{d} = P_{d'} = 1/2$ and 
$c_{\text{puct}}P_i = c$ for some constant $c > 0$.
That is,
\begin{align}\label{eq:B_alphago}
    B_{i, t+1} = X_{i,t} + c \frac{\sqrt{m_{d-1}}}{m_i+1}\,.
\end{align}
Similarly to \cref{eq:rec}, on trajectory $T+1$ where the maximum reward is reached, we can derive:
\begin{align}
    \frac{\sqrt{m_{d-1}}}{m_d+1}\geq \frac{\sqrt{m_{d-1}}}{m_{d'}+1} + \frac{1}{D}\,.
\end{align}
Still assuming that ties are broken in favour of $n_{d'}$,
we deduce that for all $d\leq D, m_{d}\leq m_{d'}$ and thus $m_{d-1}\geq 2m_d$.
Since $m_{d-1}  = m_d + m_{d'}+1$
(due to one node expansion per trajectory)
we have $m_{D-1} = 2$ 
and thus for all $d < D$,
\begin{align}
    m_{d} \geq 2^{D-d}\,.
\end{align}
From \cref{eq:B_alphago}, we also deduce that 
\begin{align}
    m_{d-1} \geq \left(\frac{m_d}{cD}\right)^2\,.
\end{align}
Similarly to \cref{eq:m0_md} where we now take $\tilde D = cD$, we obtain for all $d < D$
\begin{align}
    m_0 \geq \left(\frac{m_{d}}{c^2D^2}\right)^{2^{d}}\,.
\end{align}
Now we want to choose a $\hat d$ such that $m_{\hat d} / c^2D^2 \geq 2$,
which is satisfied for $2^{D-\hat d} \geq 2 c^2D^2$, which is satisfied for
$\hat d = \floor{D - 1 - 2\log_2(cD)}$.
This gives
\begin{align}
    T = m_0 ~\geq~ \exp_2(\exp_2(\floor{D - 1 - 2\log_2(cD)}))\,.
\end{align}
For example, for $c=2$ ($c_{\text{puct}}=4$) and $D=20$ this gives $T \geq 2^{2048}\geq 10^{200}$.

\section{UCT lower bound}

UCT uses the following formula,
for $i\in\{d,d'\}$ for all $d<D$:
\begin{align}
    B_{i, t+1} = X_{i,t} + \sqrt{\frac{2\ln m_{d-1,t}}{m_{i,t}}}\,.
\end{align}
Similarly to the previous sections, on the trajectory $T+1$ we obtain, for all $d < D$,
\begin{align}
    B_{d', T+1} &\leq B_{d', T}\,, \\[2ex]
    1-\frac{d}D + \sqrt{\frac{2\ln m_{d-1,T}}{m_{d',T}}}
    &\leq 1-\frac{d+1}D + \sqrt{\frac{2\ln m_{d-1,T}}{m_{d,T}}}\,,
    \label{eq:uct_rec}
\end{align}
from which we deduce (omitting $T$ subscripts):
\begin{align}
    \sqrt{\frac{2\ln m_{d-1}}{m_d}} &\geq \frac1D \,,\\
    m_{d-1} &\geq \exp\left(\frac{m_d}{2D^2}\right)\,.
\end{align}
Observe that starting the recursion with $m_d \leq 2D^2$ gives at most $m_{d-1} \geq \exp(1)$ which is less than $2D^2$ (for $D\geq 2$) and the recurrence actually converges to a number close to 1.
Hence, for the exponential behaviour to kick in, we need to start with $m_d$ large enough, that is, we need to find some $\hat d$ such that 
\begin{align}
    \exp\left(\frac{m_{\hat d}}{2D^2}\right) \geq 2 m_{\hat d}\,.
\end{align}
If $m_{\hat d} \geq 4D^3$ then it can be shown that the relation above is true for all $D\geq 3$.

From \cref{eq:uct_rec} we deduce that \cref{eq:mhatd} still holds,
which means that 
$m_{\hat d} \geq 4D^3$ is satisfied when $2^{D-\hat d - 1} \geq 4D^3$,
which is satisfied for
$\hat d = \floor{D - 3\log_2 D - 3}$.

We conclude that
\begin{align}
    m_0 \geq \exp(\dots \exp(\exp(4D^3 / 2D^2) / 2D^2) \dots/2D^2)\,,
\end{align}
where the number of exp is $\floor{D - 3\log_2 D - 3} = \Omega(D)$.
For $D=16$ already, $m_0 \geq e^{e^{25}}$.

\bibliographystyle{named}
\bibliography{biblio}

\end{document}